\documentclass{article}
\usepackage{nips_2018}
\usepackage[utf8]{inputenc} 
\usepackage[T1]{fontenc}   
\usepackage{hyperref}       
\usepackage{url}            
\usepackage{booktabs} 
\usepackage{amsfonts}       
\usepackage{nicefrac}       
\usepackage{microtype} 
\usepackage{color}
\usepackage{caption}
\usepackage{graphicx}
\newcommand{\sys}{AgileNet}

\usepackage{multirow}
\usepackage{natbib}
\usepackage{algorithm}
\usepackage{algpseudocode}
\usepackage{pseudocode}
\usepackage{amsmath}
\usepackage{mathtools}
\usepackage[toc,page]{appendix}
\title{AgileNet: Lightweight Dictionary-based \\ Few-shot Learning}

\begin{document}
\maketitle

\begin{abstract}
The success of deep learning models is heavily tied to the use of massive amount of labeled data and excessively long training time. With the emergence of intelligent edge applications that use these models, the critical challenge is to obtain the same inference capability on a resource-constrained device while providing adaptability to cope with the dynamic changes in the data.
We propose \sys{}, a novel lightweight dictionary-based few-shot learning methodology which provides reduced complexity deep neural network for efficient execution at the edge while enabling low-cost updates to capture the dynamics of the new data. Evaluations of state-of-the-art few-shot learning benchmarks demonstrate the superior accuracy of \sys{} compared to prior arts. Additionally, \sys{} is the first few-shot learning approach that prevents model updates by eliminating the knowledge obtained from the primary training. This property is ensured through the dictionaries learned by our novel end-to-end structured decomposition, which also reduces the memory footprint and computation complexity to match the edge device constraints.
\end{abstract}

\section{Introduction} 
Deep Neural Networks (DNNs) have achieved a remarkable success in several critical application domains including computer vision, speech recognition, and natural language processing. The trend of making deeper and wider networks to achieve higher model accuracy counters the goal of providing networks with higher efficiency in terms of model size and the speed of training/inference. Efficiency and compactness are of growing concerns since many of the applications relying on deep learning models are eventually aimed at providing intelligence on resource-constrained devices at the edge. The conventional cloud outsourcing approach fails to address latency, privacy, and availability concerns~\cite{howard2017mobilenets, abadi2016deep}. This has been the catalyst for a large number of works building efficient DNN inference accelerators such as~\cite{lane2015early, dnnweaver:micro16}. Training phase of DNNs incurs a larger memory footprint and computation complexity compared with the inference. Assuming training is a one-time task, after which the model can be deployed on the inference accelerator platform on the edge, the major trend has been to train on the cloud~\cite{lane2016deepx}. However, providing adaptability at the edge is necessary to maintain the desired accuracy in dynamic environment settings. 

To address the above requirements, there is a need to tackle two key challenges so they can effectively fit within the edge devices: (i) how to reduce memory and computation cost of the DNN model on the cloud server without compromising the application performance and accuracy. (ii) how to extend the space of model parameters to learn new tasks on-device without forgetting the knowledge learned originally. Learning new tasks should be performed using few data instances over few iterations to comply with the stringent physical performance requirements at the edge.

Conventional supervised deep learning is dependent on the availability of a massive amount of labeled data; the trained models generally perform poorly when labeled data is limited. The problem of rapidly learning new tasks with a limited amount of labeled data is referred to as ``few-shot learning'', which has received considerable attention from research community in recent years~\cite{fei2006one, lake2015human, hariharan2017low}. However, many of the recent approaches solely consider the model's performance on the new task and thus their approach discards the primary knowledge of the older tasks. This is in contrast with the goal of providing adaptable intelligence at the edge where adding to the capabilities of the model is desired without forgetting the previous knowledge. Neglecting the physical constraints of the edge device, in terms of memory, compute power, and energy consumption is another drawback of many of the state-of-the-art few-shot learning approaches. A practical few-shot learning methodology should extend the capabilities of the model not only using the few available new data instances but also through lightweight updates to the model.

This work proposes \sys{}, the first lightweight few-shot learning scheme that enables efficient and adaptable edge device realization of DNNs. To enable \sys{}, we create a novel end-to-end structured decomposition methodology for DNNs which allows low-cost model updates to capture the dynamics of the new data. \sys{} not only performs lightweight and effective few-shot leaning but also shrinks the storage requirement and computational cost of the model to match the edge device constraints. In summary, the contributions of this work are as follows:

\begin{itemize}
    \item Proposing \sys{}, a novel dictionary-based few-shot learning approach to enable adaptability at the edge while complying with the stringent resource constraints.
    \item Developing a new end-to-end structured decomposition methodology which reduces memory footprint and computational complexity of the model to match edge constraints.
    \item Innovating a lightweight model updating mechanism to capture the dynamics of the new data with only a few instances leveraging the properties of the learned dictionaries.   
   \item Demonstrating the superior accuracy of \sys{} on few-shot learning benchmarks compared with the state-of-the-art approaches on standard few-shot learning benchmarks. \sys{} is shown to preserve the accuracy on old and new classes, while reducing the amount of storage and computing.
\end{itemize}

The rest of paper is structured as follows. Section~\ref{sec:related} provides a review of related literature and discusses drawbacks of the prior art. The global flow of \sys{} is described in Section~\ref{sec:global_flow}. Section~\ref{sec:methodology} presents the details of the structured decomposition methodology. Few-shot learning technique is explained in Section~\ref{sec:fewshot}. Section~\ref{sec:evaluation} provides the experiment setting and benchmark evaluations and is followed by conclusions in Section~\ref{sec:conclusion}.

\section{Related Work}\label{sec:related}

The key challenge of few-shot learning is to use primary knowledge obtained through original training data to make predictions about unseen classes of data with a limited number of available samples. Following the long history of research on few-shot learning approaches, the first work to leverage modern machine learning for one-shot learning was proposed by ~\cite{fei2006one}. In recent years, the work in ~\cite{lake2015human} and ~\cite{koch2015siamese} have established two standard benchmarks, Omniglot and Mini-ImageNet respectively, to compare few-shot learning approaches in terms of accuracy.  ~\cite{lake2015human} leverages a Bayesian model while the authors of ~\cite{koch2015siamese} utilized a Siamese network which learns pairwise similarity metrics to generalize the predictive power of the model to new classes. These works were followed by other pairwise similarity-based few-shot learning approaches in ~\cite{vinyals2016matching,snell2017prototypical,mehrotra2017generative}. 

From a different perspective, few-shot learning through combining graph-based analytics with deep learning has been proposed in~\cite{garcia2017few}. In a separate trend of work, meta-learners~\cite{ravi2016optimization,munkhdalai2017meta,mishra2017simple} are developed to generalize the DNN model to new related tasks. The aforementioned works have incrementally increased the accuracy on few-shot learning benchmarks. However, all these works are negligent to the model accuracy on old classes. Therefore, their proposal can degrade the predictive power of the model on old data. Additionally, many of the aforementioned approaches incur a high computation cost to adapt the model and thus are not amenable to resource-constrained settings. \sys{} preserves the prior knowledge of the model on old data while outperforming all state-of-the-art approaches in terms of few-shot learning accuracy. Additionally, lightweight model updates of \sys{} complies with stringent limitations of edge devices.

\section{Global Flow of \sys{}} \label{sec:global_flow}

\begin{figure}[h!]
\begin{center}
  \includegraphics[width=0.77\textwidth]{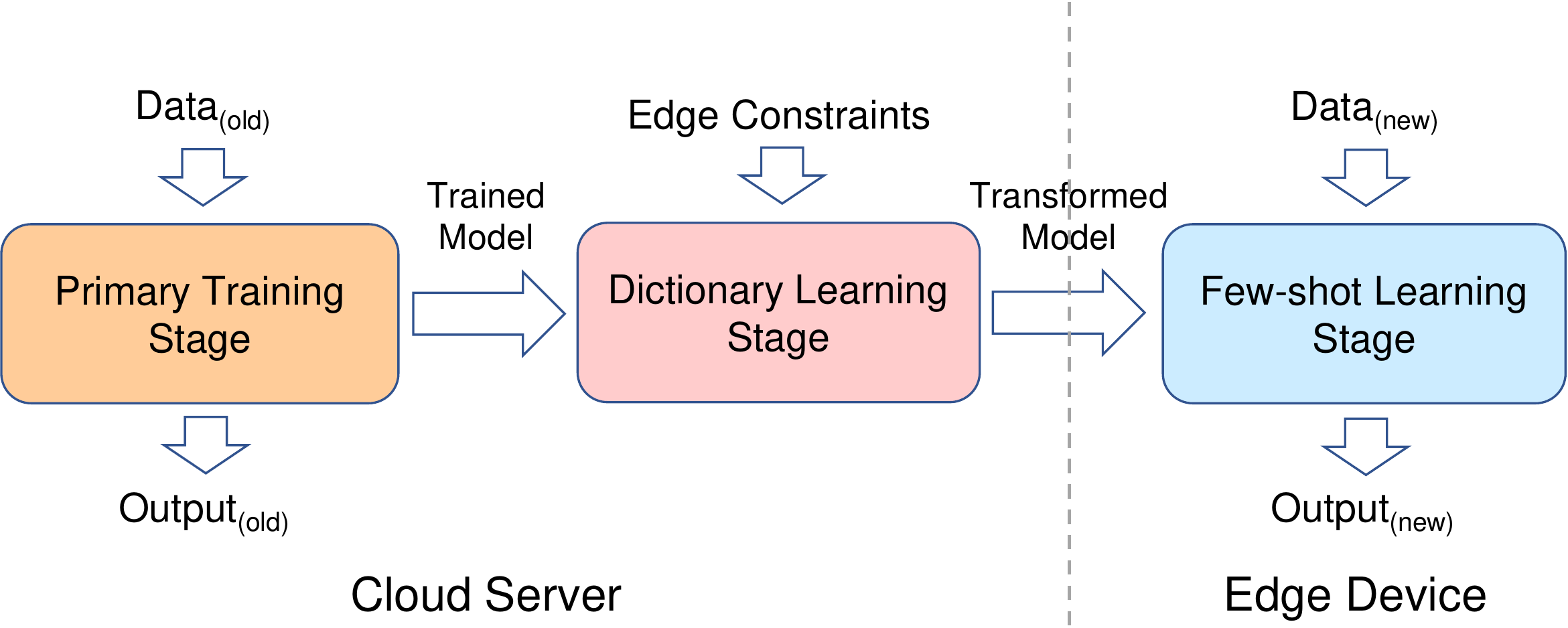}
  \caption{Global flow of \sys{}}
  \label{fig:flobal_flow}
\end{center}
\end{figure}

Figure~\ref{fig:flobal_flow} presents the global flow of \sys{} which involves three stages: primary training stage, dictionary learning stage, and few-shot learning stage.  The first two stages are performed on the cloud, and the last stage is executed on the edge device with limited resources.

\noindent \textbf{Primary Training Stage:} At this stage, the original model with the mainstream architecture is trained using conventional training methodologies.

\noindent \textbf{Dictionary Learning Stage:} The trained model and edge constraints in terms of the memory and computation resources are taken into account for transforming the model using end-to-end structured decomposition discussed in Section~\ref{sec:methodology}. At this stage, the trade-off between memory/computation cost and final accuracy of the model is leveraged to match the edge constraints. 


\noindent \textbf{Few-shot Learning Stage:} Finally, the \sys{} model is deployed on the edge device. Despite the memory and computational benefits of structured decomposition, it enables adaptability for dynamic settings. The~\sys{} model provides the expected inference accuracy on the desired task under tight resource constraints. At the same time, when encountering new classes of data, low-cost updates on the edge device are sufficient to learn new capabilities.

\section{End-to-End Structured Decomposition} \label{sec:methodology}
\sys{} performs structured decomposition on all layers using an adaptive subspace projection method built on the foundation of column subset selection proposed in~\cite{tropp2009column, boutsidis2009improved}. We emphasize that \sys{} is the first to leverage this technique to perform an end-to-end transformation of a DNN model; however, the work in ~\cite{rouhani2016delight} used a similar approach to project input data to a DNN model into lower dimensions.

\subsection{Adaptive Subspace Projection}
Assume an arbitrary matrix $W_{m\times n}$. The goal of subspace projection technique is to represent $W_{m\times n}$ with a coefficient matrix $C_{l\times n}$ and a basis dictionary matrix $D_{m\times l}$ such that $l<<n$ and $|W - DC| < \beta$,
where $l$ is dimensionality of the ambient space after projection and $\beta$ is the absolute tolerable error threshold for the projection. This decomposition allows us to represent matrix $W_{m\times n}$ with correlated columns using the coefficients $C_{l\times n}$ and the dictionary $D_{m\times l}$ with negligible error. To build coefficient and dictionary matrices, adaptive subspace projection adds a particular column of $W_{m\times n}$ that minimizes the projection error to the dictionary matrix at each iteration. According to the desired error threshold, this technique creates the dictionary by increasing $l$, which is the number of columns of dictionary $D$ until it finds a suitable lower-dimensional subspace for the data projection. This dictionary can be adaptively updated as the dynamics of the original $W$ matrix change by appending new columns to it.



\subsection{Layer-wise Dictionary Learning}\label{sec:layerdiction}
Neural network computations are dominated by matrix multiplications. At dictionary learning stage, trained DNN weight matrix for each layer is decomposed into a dictionary matrix and a coefficient matrix according to an error threshold $\beta$ which can be adjusted per layer. Next, we will explain this structured decomposition for fully-connected (fc) and convolution layers (conv).

\begin{figure}[h!] 
\begin{center}
  \includegraphics[width=0.7\textwidth]{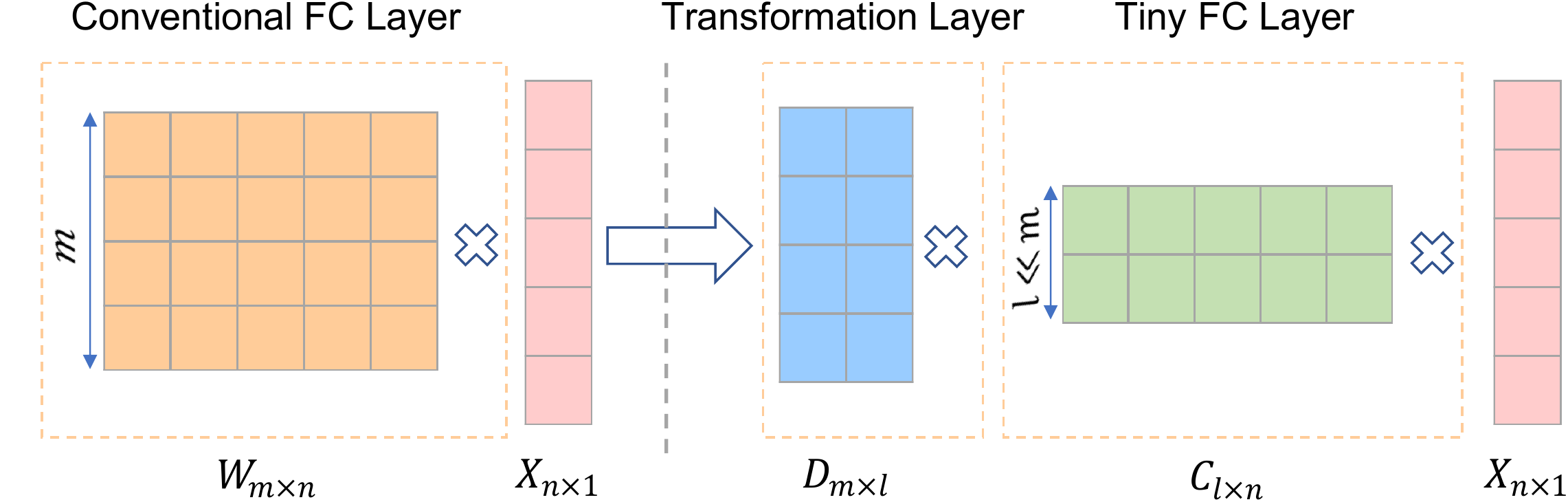}
  \vspace{0.5em}
  \caption{Transformation of fully-connected layer in \sys{}.}\label{fig:trans_fc}
  \end{center}
\end{figure}

\noindent \textbf{Fully-connected layer:} In a conventional fully-connected layer, the following matrix-vector multiplication is performed
\begin{equation}{y_{m\times 1} = W_{m\times n}x_{n\times 1},}\label{eq:fc}
\end{equation}
where $x$ and $y$ are input and output vectors respectively. In our scheme, weight matrix $W$ is transformed into dictionary matrix $D$ and coefficient matrix $C$. Substituting this into Equation~\ref{eq:fc} results in:
\begin{equation}{y_{m\times 1} = D_{m\times l}C_{l\times n}x_{n\times 1},}\label{eq:fc_decomposed}
\end{equation}
In \sys{}, the above equation is performed by two subsequent layers. In particular, a conventional fully-connected layer is replaced by a tiny fully-connected layer (with weight matrix $C$) preceded by an transformation layer (with weight matrix $D$) as shown in Figure~\ref{fig:trans_fc}.

\begin{figure}[h!]
  \begin{center}
  \includegraphics[width=0.9\textwidth]{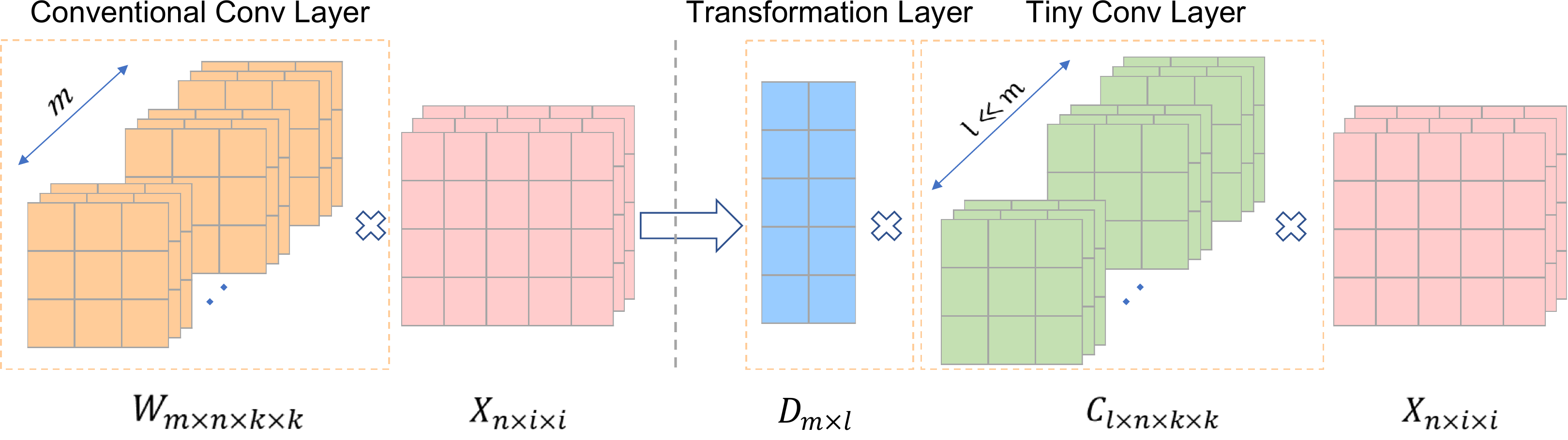}
  \vspace{1em}
  \caption{Transformation of convolution layer in \sys{}.}
  \label{fig:trans_conv}
  \end{center}
\end{figure}

\noindent \textbf{Convolution Layer:}
For a convolution layer, we first matricize weight tensor $W$. After subspace projection, dictionary $D$ remains intact while the coefficient matrix $C$ is reshaped into a three-dimensional tensor. The reason for this decision is to comply a universal format for the dictionaries in all layers. Similar to a fully-connected layer, substituting the weight tensor $W$ of a convolution layer with dictionary matrix $D$ and coefficients tensor $C$ transforms a conventional convolution (with $m$ output channels) into a tiny convolution layer (with $l<<m$ output channels) preceded by a transformation layer as shown in Figure~\ref{fig:trans_conv}. For any row of $D$, each element is multiplied by all elements of the corresponding channel (of the tiny conv layer output) and resulting channels are summed up element-wise to generate one output channel. As such, the transformation layer takes an $l$-channel input and transforms it into an $m$-channel output using a linear combination approach.

\subsection{End-to-End Dictionary Learning}\label{sec:end2end}
At dictionary learning stage, weights of the trained model are initially decomposed into a dictionary matrix and a coefficient matrix to comply with the edge constraints.  The transformed model has the same architecture of layers as the original model but the fully-connected and convolution layers are replaced by their corresponding transformation layer and tiny layer as discussed above. To compensate for possible loss of accuracy as the result of the structured decomposition, the transformed model is fine-tuned. Note that very tight memory/compute budget at the edge might result in a transformed model that is not inherently capable of achieving the desired accuracy. Additionally, we empirically realized that last few fully-connected layers in a DNN architecture contribute more to the final model accuracy and therefore require a smaller decomposition error threshold. These two important observations are explored in Section~\ref{sec:evaluation}. At the end of this stage, the transformed model of \sys{} can be readily deployed on the edge device.

\section{Few-shot Learning on the Edge Device}\label{sec:fewshot}
The stringent memory and energy constraints at the edge are the major challenges towards on-device training of the neural networks. This limitation is due to the power-hungry computations of the training phase as well as excessive memory requirement of large models used in real-world applications. The prohibitive memory cost of primary training data hinders its storage on the edge device to be used for model updating. The new data is also available only in few instances. Also, adapting the model to new data should not exacerbate the performance on old classes.

Structured decomposition generates dictionaries that preserve the structure of weights in each layer and are built such that they capture the space of weight parameters. We leverage this property for updating the model in few-shot learning scenarios: \sys{} keeps the dictionary for all layers intact and only fine-tunes the coefficients. A minute update to the coefficients of \sys{} should be enough to expand the capability of the model for inference on new data. This means that the model can be tuned for new data through only a small number of iterations. Additionally, since the coefficient matrix (tensor) is considerably smaller than the original weight matrix (tensor), a smaller number of parameters need to be updated for \sys{}. In particular, the number of trainable parameters for few-shot learning tasks is reduced by approximately a factor $\frac{l}{m}$ for both fully-connected and convolution layers, where $l<<m$ is the number of rows (channels) in the coefficient matrix (tensor) and $m$ is the number of rows (channels) in the original weight matrix (tensor). Note that as we show in Section~\ref{sec:evaluation}, our approach for few-shot learning also preserve the predictive power of the model on original classes.
To enable on-device training under more strict compute/energy budgets, we introduce an ultra-light mode which reduces parameter updates even more.

\begin{figure}[h!] 
\begin{center}
  \includegraphics[width=0.8\textwidth]{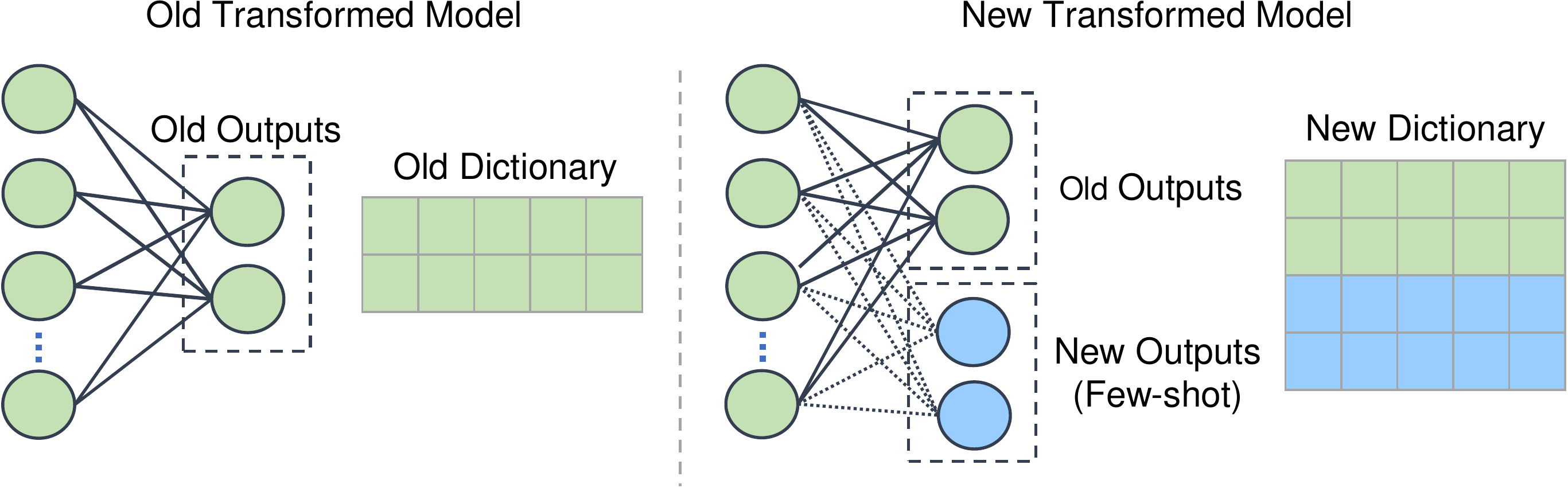}
  \caption{Ultra-light few-shot learning mode of \sys{}. For this example, only two rows of the new dictionary matrix (for the last layer, corresponding to the new data classes) are being updated.}\label{fig:ultralight_mode}
  \end{center}
\end{figure}

\noindent \textbf{Ultra-Light Few-shot Learning:}
This mode is designed to further limit the cost of model adaptation at the edge, though it might also limit the maximum achievable accuracy on the new data. In ultra-light few-shot learning mode, all layers except the last layer are not trained; not only the dictionaries but also the coefficients of all layers except the last remain intact. Furthermore, the coefficient of the last fully-connected layer, as well as rows of its dictionary matrix that correspond to old data classes, are fixed. The only parameters that are updated belong to the few rows of the dictionary matrix that correspond to the new data categories. This mode, which is depicted in Figure~\ref{fig:ultralight_mode}, has significantly fewer parameters to fine-tune and thus, converges in a much small number of iterations.

\section{Experiments and Evaluation} \label{sec:evaluation}
Our evaluation are performed on three benchmark datasets: MNIST~\cite{lecun2010mnist}, Omniglot~\cite{lake2015human} and Mini-Imagenet~\cite{vinyals2016matching}. $X-way$ $Y-shot$ learning experiment is performed as follows: we randomly sample $X$ classes from the test dataset. From each selected class, we choose $Y$ data instances randomly. We feed the corresponding $X\times Y$ labeled examples to the model during the few-shot learning stage. The trained model is then tested on the data from the same $X$ classes excluding $Y$ examples used for few-shot learning. The top-1 average test accuracy is reported for different random new classes and different data instances within each new class. Note that for all experiments, we followed all three steps of the global flow of \sys{}.

\noindent \textbf{MNIST:}
The dataset of handwritten digits $0$ to $9$ consists of $60,000$ examples in the training set, and $10,000$ examples in the test set of size $28\times 28$. For few-shot learning experiments, we randomly chose $9$ digits for primary training and the remaining digit was used as the new classes in the few-shot setting. We used LeNet architecture which has two convolution layers with the kernel size of $5\times 5$, followed by a dropout layer, and two fully-connected layers. To validate \sys{} in few-shot learning scenarios, we chose five samples from all ten classes randomly and created a new training set for few-shot learning stage. Since this data contains five data instances from the new class, the few-shot task is 1-way 5-shot learning. We note that adding samples from old classes to the training data for few-shot learning stage is to preserve model accuracy on old classes and prevent over-fitting on the new class. However, we only need to store 5 samples of each old class on the edge device for this purpose which does not add significant memory overhead to this stage.

\begin{figure}[h!] 
\begin{center}
  \vspace{-2mm}
  \includegraphics[width=\textwidth]{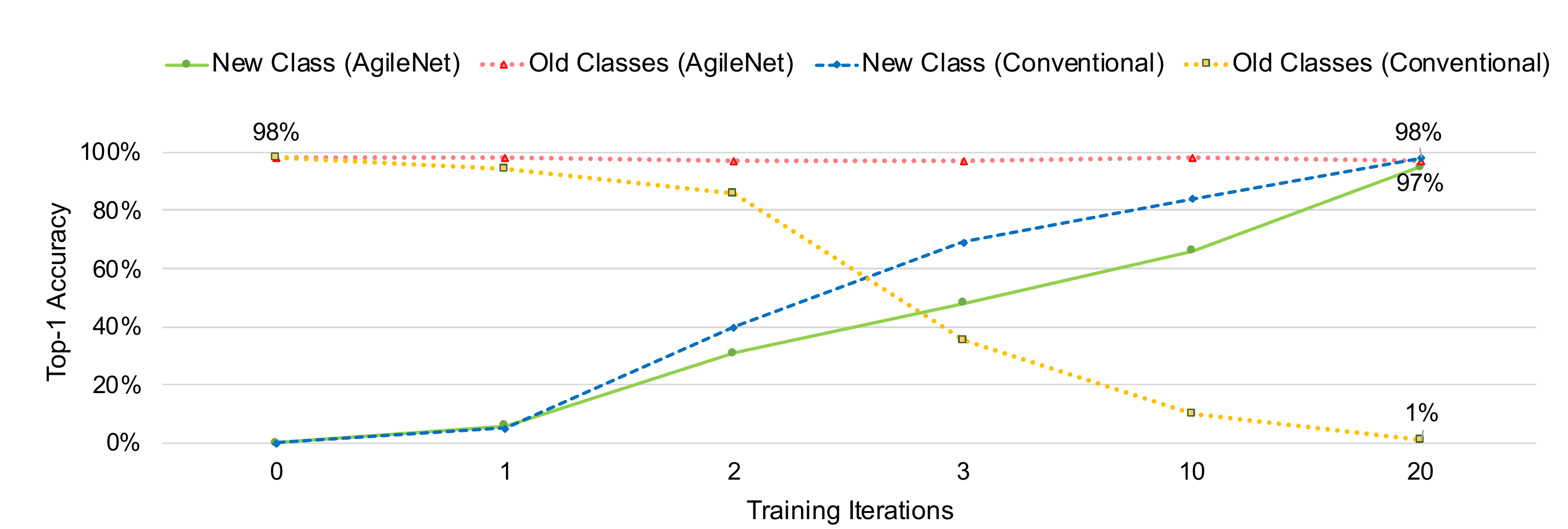}
   \vspace{0mm}
  \caption{Classification accuracy on the test set during the few-shot learning stage (1-way 5-shot) for \sys{} and conventional training method.} \label{fig:mnist}
  \end{center}
\end{figure}

Figure~\ref{fig:mnist} shows the classification accuracy on the test set after few-shot learning stage. The green and red lines represent \sys{} accuracy on the new and old classes, respectively. Our approach achieves a reasonable accuracy of $97\%$ only after 20 iterations while preserving $98\%$ accuracy on original 9 classes. However, conventional training sacrifices the accuracy on the old classes to obtain a comparable accuracy to \sys{} on the new class. There are two key factors for success of \sys{} in preserving the knowledge on old classes: (i) The learned dictionaries preserve the structure of the weights at each layer and minute coefficient updates do not exacerbate the accuracy on the old classes. (ii) Our training data covers samples from both old and new classes to prevent over-fitting.


\noindent \textbf{Omniglot:}
This benchmark dataset for few-shot learning tasks has 50 different alphabets including 1623 character classes totally. Each character class has only $20$ samples. The dataset is split into the first 1200 classes for training and the remaining 423 classes for testing as in~\cite{vinyals2016matching,garcia2017few}. Images are resized to $28\times 28$. To compensate for the low number of training examples per class, we appended three rotated images (by 90, 180, and 270 degrees) for each original image to the training data.  

For this dataset, we used the CNN architecture proposed by ~\cite{vinyals2016matching} in which each block has a convolution layer with 64 filters of size $3\times 3$, a batch-normalization layer, a $2\times 2$ max-pooling layer, and a leaky-relu. Four of these blocks are stacked and followed by a final fully-connected layer. We experimented $5-way$ $1-shot$, $5-way$ $5-shot$, $20-way$ $1-shot$, and $20-way$ $5-shot$ scenarios. Table~\ref{tab:table_omiglot} compares final accuracy of \sys{} with prior work on these tasks. For $5-way$ experiments, \sys{} outperforms all prior work and for $20-way$ tasks, it achieves a comparable accuracy. Note that for these experiments, we are only comparing the accuracy on the new classes as all of the prior works in Table~\ref{tab:table_omiglot} have used this metric for comparison. As such, the training data for few-shot learning stage consists of only samples from the new class.

\begin{table}
\centering
\begin{tabular}{l|cc|cc}
\hline
\multirow{2}{*}{Model}                       & \multicolumn{2}{c|}{5-Way} & \multicolumn{2}{c}{20-Way} \\
                                            & 1-shot       & 5-shot      & 1-shot       & 5-shot      \\ \hline
Matching Networks~\cite{vinyals2016matching}     & 98.1\%       & 98.9\%      & 93.8\%       & 98.5\%      \\
Statistic Networks~\cite{edwards2016towards}   & 98.1\%       & 99.5\%      & 93.2\%       & 98.1\%      \\
Res. Pair-Wise~\cite{mehrotra2017generative} & -            & -           & 94.8\%       & -           \\
Prototypical Networks~\cite{snell2017prototypical}   & 97.4\%       & 99.3\%      & 95.4\%       & 98.8\%      \\
ConvNet with Memory~\cite{DBLP:journals/corr/KaiserNRB17}    & 98.4\%       & 99.6\%      & 95.0\%       & 98.6\%      \\
Agnostic Meta-learner~\cite{finn2017model}    & 98.7\%       & \textbf{99.9}\%      & 95.8\%       & 98.9\%      \\
Meta Networks~\cite{munkhdalai2017meta}       & 98.9\%       & -           & 97.0\%       & -           \\
TCML~\cite{mishra2017meta}                  & 98.96\%      & 99.75\%     & \textbf{97.64}\%      & \textbf{99.36}\%     \\
GNN~\cite{garcia2017few}                  & 99.2\%       & 99.7\%      & 97.4\%       & 99.0\%      \\ \hline
\sys{} (Ours)                                    & \textbf{99.5}\%       & \textbf{99.9}\%      & 94.95\%      & 98.9\%      \\ \hline
\end{tabular}
\vspace{4mm}
\caption{Comparison of classification accuracy after few-shot learning on Omniglot dataset with 95\% confidence intervals. The best results of each scenario are marked in bold.}
\vspace{-4mm}
\label{tab:table_omiglot}
\end{table}

\noindent \textbf{Mini-Imagenet:}
A more challenging benchmark for few-shot learning experiments was proposed by~\cite{vinyals2016matching} which is extracted from the original Imagenet dataset. Mini-Imagenet consists of 60,000 images of size $84\times 84$ belonging to 100 classes. We used first 64 classes for training, 16 classes for validation and last 20 for test similar to~\cite{ravi2016optimization}. The CNN architecture used in this experiment consists of 4 convolution layers. Each convolution layer has a different number of filters (64, 96, 128, 256) with the kernel size of $3\times 3$ followed by a batch normalization layer, a max pooling layer, and a leaky-relu. The last two convolution layers are also followed by a dropout layer to avoid over-fitting. This architecture has a fully-connected layer at the end.

\begin{figure}[h!] 
\begin{center}
  \includegraphics[width=0.85\textwidth]{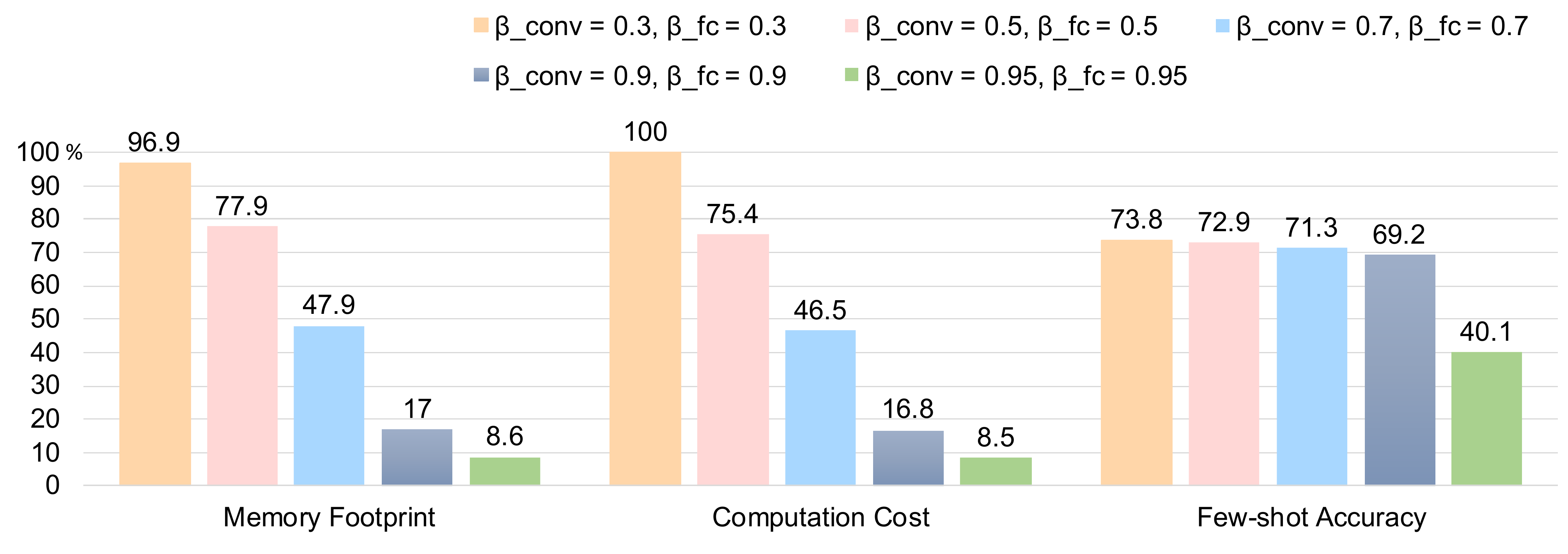}
  \caption{Comparison of memory footprint, computation cost and final few-shot learning accuracy ($5$-way $5$-shot task) of \sys{} with different decomposition error thresholds.} \label{fig:epsilon_experiment}
  \end{center}
\end{figure}

In order to explore the space of decomposition error threshold $\beta$ for different layers which determines the dictionary size (and in turn, memory footprint and computation cost) as well as the model accuracy, we conducted a comprehensive analysis of \sys{} for Mini-Imagenet dataset. Figure~\ref{fig:epsilon_experiment} presents the trade-off between memory footprint, computation cost and final accuracy after few-shot learning stage corresponding to different decomposition error thresholds $\beta$ uniformly set for all layers. Memory and computation costs are compared with those of the original model in the primary training stage and the few-shot learning accuracies denote the absolute test accuracy on new classes. Notice that memory footprint and computation cost decrease significantly as $\beta$ increases. The drop in accuracy of \sys{} is negligible until $\beta$ reaches $0.95$. Then, the model accuracy drops to $40.1$\%. These results demonstrate that the trade-off between model accuracy and memory/computation cost can be leveraged by adjusting the decomposition error threshold. This flexibility allows \sys{} to match the edge device physical constraints while enabling the desired degree of adaptability to new data. 

\begin{figure}[h!] 
\begin{center}
  \vspace{-3mm}
  \includegraphics[width=0.85\textwidth]{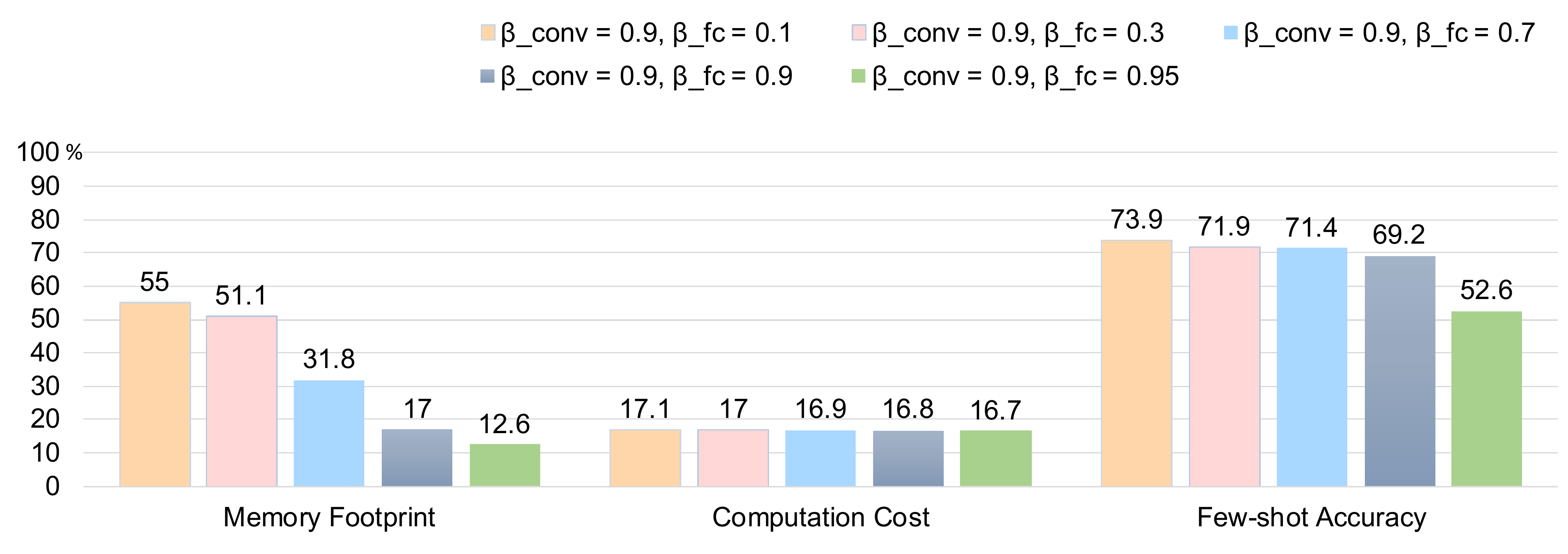}
  \caption{Comparison of memory footprint, computation cost and few-shot learning accuracy ($5$-way $5$-shot task) with a varying decomposition error threshold only for the last layer. The light and blue bars reduce memory footprint by 3.1$\times$ and 5.8$\times$, respectively. Computation cost for all these configurations is approximately 6$\times$ less than the original model.} \label{fig:epsilon_experiment_fc}
  \end{center}
\end{figure}

To further understand the impact of decomposition error threshold on convolution layers and fully-connected layers, we varied $\beta$ for the fully-connected layer in this DNN architecture from $0.1$ to $0.95$ while keeping $\beta$ for all convolution layers intact as shown in Figure~\ref{fig:epsilon_experiment_fc}. Changing $\beta$ for the fully-connected layer mainly impacts the memory footprint of the model while the computation cost is dominated by the convolution layers. Similar to the previous experiment, memory footprint decreases as $\beta$ increases for the fully-connected layer. The drop in accuracy of \sys{} is negligible until $\beta$ reaches $0.95$. Then, the model accuracy drops to $52.6$\%. These results show that layer-wise exploration of the decomposition error threshold is necessary to maximize the memory/computation benefits of \sys{} while achieving a desired accuracy for few-shot learning tasks. 

To compare \sys{} with prior work, we used two configurations for decomposition error thresholds for different layers as shown in Table~\ref{tab:table_imagenet}. In both $1$-shot and $5$-shot scenarios, \sys{} achieves a higher accuracy than all prior arts. Similar to the Omniglot benchmark, these results only consider the accuracy on new classes. We emphasize that \sys{} outperforms prior works in terms of accuracy while reducing memory footprint by \textbf{5.8}$\times$ and computation cost by \textbf{6}$\times$ as shown in Figure~\ref{fig:epsilon_experiment_fc}. Therefore, \sys{} not only helps amenability of large DNN models to resource-constrained devices but also it achieves a superior accuracy in few-shot learning scenarios compared to the state-of-the-art.

\begin{table}[h]
\centering
\vspace{5mm}
\begin{tabular}{l|cc}
\hline
\multirow{2}{*}{Model}                         & \multicolumn{2}{c}{5-Way}                                                                                                                          \\
                                               & 1-shot                                                                   & 5-shot                                                                   \\ \hline
Matching Networks~\cite{vinyals2016matching}        & 43.6\%                                                                   & 55.3\%                                                                   \\
Prototypical Networks~\cite{snell2017prototypical}     & 46.61\% $\pm$ 0.78\%                                                     & 65.77\% $\pm$ 0.70\%                                                     \\
Model Agnostic Meta-learner~\cite{finn2017model} & 48.70\% $\pm$ 1.84\%                                                     & 63.10\% $\pm$ 0.92\%                                                      \\
Meta Networks~\cite{munkhdalai2017meta}          & 49.21\% $\pm$ 0.96\%                                                     & -                                                                        \\
M. Optimization~\cite{ravi2016optimization}   & 43.40\% $\pm$ 0.77\%                                                      & 60.20\% $\pm$ 0.71\%                                                      \\
TCML~\cite{mishra2017meta}                      & 55.71\% $\pm$ 0.99\%                                                     & 68.88\% $\pm$ 0.92\%                                                     \\
GNN~\cite{garcia2017few}                     & 50.33\% $\pm$ 0.36\%                                                     & 66.41\% $\pm$ 0.63\%                                                     \\ \hline
\sys{} ($\beta_{conv}=0.9, \beta_{fc}=0.9$)                                    & \textbf{48.38}\% $\pm$ \textbf{0.90}\% & \textbf{69.21}\% $\pm$ \textbf{0.25}\% \\
\sys{} ($\beta_{conv}=0.9, \beta_{fc}=0.7$)                                    & \textbf{58.23}\% $\pm$ \textbf{0.10}\% & \textbf{71.39}\% $\pm$ \textbf{0.10}\% \\
\hline
\end{tabular}
\vspace{5mm}
\caption{Comparison of classification accuracy after few-shot learning on Mini-Imagenet dataset with 95\% confidence intervals. The best results of each scenario are marked in bold.}
\label{tab:table_imagenet}
\end{table}

\section{Conclusion}\label{sec:conclusion}
This work presents the first lightweight few-shot learning approach that beats the accuracy of state-of-the-art approaches on standard benchmarks through only a small number of parameter updates. The key enabler of \sys{} is our novel end-to-end structured decomposition methodology that replaces every convolution and fully-connected layer by its tiny counterpart such that memory footprint and computational complexity of the transformed model matches the edge constraints. Our experiments corroborated that the learned dictionaries of \sys{} preserve the structure of the model, enabling low-cost and effective few-shot learning without degrading the model accuracy on old data classes.
\bibliographystyle{plainnat} 

\small
\bibliography{mybibliography}

\end{document}